\documentclass[10pt,twocolumn,letterpaper]{article}

\usepackage{colortbl}
\usepackage{cvpr}
\usepackage{times}
\usepackage{epsfig}
\usepackage{graphicx}
\usepackage{amsmath}
\usepackage{amssymb}

\usepackage{microtype}

\usepackage{booktabs}
\usepackage[table]{xcolor}

\usepackage{multirow}

\usepackage[breaklinks=true,letterpaper=true,colorlinks,bookmarks=false]{hyperref}

\cvprfinalcopy 

\def\cvprPaperID{3680} 

\ifcvprfinal\fi

\definecolor{yelloworange}{RGB}{255, 153, 0}
\definecolor{ultramarineblue}{RGB}{65, 102, 245}

\newtheorem{lemma-ap}{Lemma}

\newtheorem{claim-ap}{Claim}

\makeatletter
\usepackage{xspace}
\def\@onedot{\ifx\@let@token.\else.\null\fi\xspace}
\DeclareRobustCommand\onedot{\futurelet\@let@token\@onedot}

\newcommand{\figref}[1]{Fig\onedot~\ref{#1}}

\newcommand{\tabref}[1]{Tab\onedot~\ref{#1}}

\def\eg{\emph{e.g}\onedot} 
\def\ie{\emph{i.e}\onedot} 
\def\cf{\emph{cf}\onedot} 
 
\def\wrt{w.r.t\onedot} 
\def\etal{\emph{et al}\onedot}

\newcommand{\eat}[1]{}

\makeatother

\begin{document}

\title{MaskLab: Instance Segmentation by Refining Object Detection with Semantic and Direction Features}

\author{
  \hspace{-1.3cm}
  \begin{tabular}[t]{c}
    Liang-Chieh Chen$^1$, Alexander Hermans$^{2*}$, George Papandreou$^1$, Florian Schroff$^1$, Peng Wang$^{3*}$, Hartwig Adam$^1$ \\
    Google Inc.$^1$, RWTH Aachen University$^2$, UCLA$^3$ \\
\end{tabular}
}

\maketitle

\ifcvprfinal
\footnotetext{$^*$Work done in part during an internship at Google Inc.}
\fi

\begin{abstract}
  In this work, we tackle the problem of instance segmentation,
the task of simultaneously solving object detection and semantic
segmentation. Towards this goal, we present a model, called MaskLab,
which produces three outputs: box detection, semantic segmentation,
and direction prediction. Building on top of the Faster-RCNN object
detector, the predicted boxes provide accurate localization
of object instances. Within each region of interest, MaskLab 
performs foreground/background segmentation by combining semantic
and direction prediction. Semantic segmentation assists
the model in distinguishing between objects of different semantic classes
including background, while the direction prediction, estimating
each pixel's direction towards its corresponding center, allows
separating instances of the same semantic class. Moreover, we
explore the effect of incorporating recent successful methods from both
segmentation and detection (\eg, atrous convolution and hypercolumn).
Our proposed model is evaluated on the COCO instance segmentation
benchmark and shows comparable performance with other state-of-art models.
\end{abstract}

\section{Introduction}
Deep Convolutional Neural Networks (ConvNets) \cite{LeCun1998, krizhevsky2012imagenet} have significantly
improved the performance of computer vision systems. In particular, models based on Fully Convolutional
Networks (FCNs) \cite{sermanet2013overfeat, long2014fully} achieve remarkable results in object detection (localize instances)
\cite{erhan2014scalable, szegedy2014scalable, girshick2015fast, ren2015faster, liu2015ssd, redmon2016you, dai2016rfcn, lin2016feature}
and semantic segmentation (identify semantic class of each pixel) \cite{chen2014semantic, lin2015efficient, papandreou2015weakly, liu2015semantic, zheng2015conditional, wu2016wider, zhao2017pyramid, luo2017deep}.
Recently, the community has been tackling the more challenging instance segmentation task \cite{girshick2014rcnn, hariharan2014simultaneous},
whose goal is to localize object instances with pixel-level accuracy, jointly solving object detection and semantic segmentation.

Due to the intricate nature of instance segmentation, one could develop a system focusing on instance
box-level detection first and then refining the prediction to obtain more detailed mask segmentation,
or conversely, one could target at sharp segmentation results before tackling the association problem
of assigning pixel predictions to instances. The state-of-art instance segmentation model FCIS \cite{dai2017fully}
employs the position-sensitive \cite{dai2016instancesensitive} inside/outside score maps to encode the
foreground/background segmentation information. 
The usage of inside/outside score maps successfully segments foreground/background regions within each
predicted bounding box, but it also doubles the number of output channels because of the redundancy of
background encoding.  On the other hand, the prior work of \cite{uhrig2016pixel} produces three
outputs: semantic segmentation, instance center direction (predicting pixel's direction towards its
corresponding instance center), and depth estimation. However, complicate template matching is employed
subsequently to decode the predicted direction for instance detection. In this work, we present
MaskLab (short for Mask Labeling), seeking to combine the best from both detection-based and
segmentation-based methods for solving instance segmentation.

\begin{figure}[!t]
  \centering
  \begin{tabular}{c c}
    \includegraphics[width=0.47\linewidth]{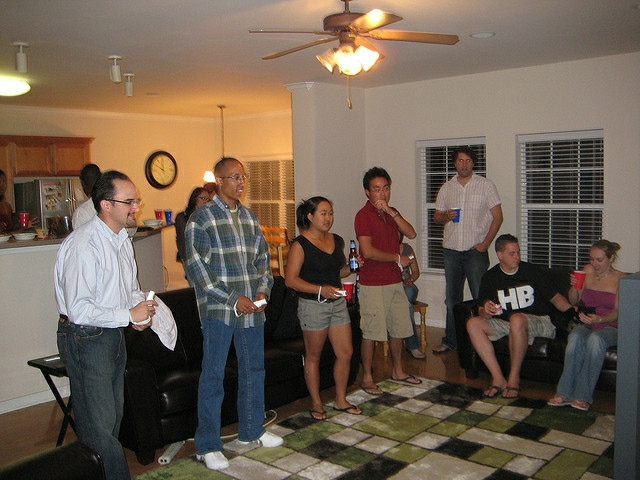} &
    \includegraphics[width=0.47\linewidth]{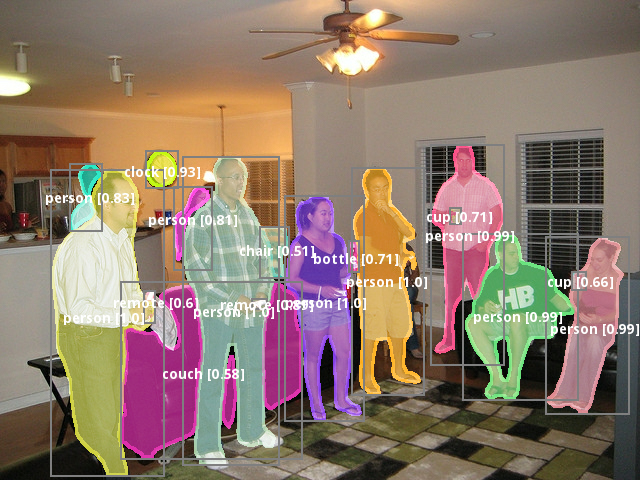} \\
    {\small(a) Image} &
    {\small(b) Predicted masks} \\
  \end{tabular}
  \caption{Instance segmentation aims to solve detection and segmentation jointly.
    We tackle this problem by refining the segmentation masks within predicted boxes (gray bounding boxes).}
  \label{fig:illustration}
\end{figure}

\begin{figure*}[!t]
  \centering
  \begin{tabular}{c}
    \includegraphics[width=0.95\linewidth]{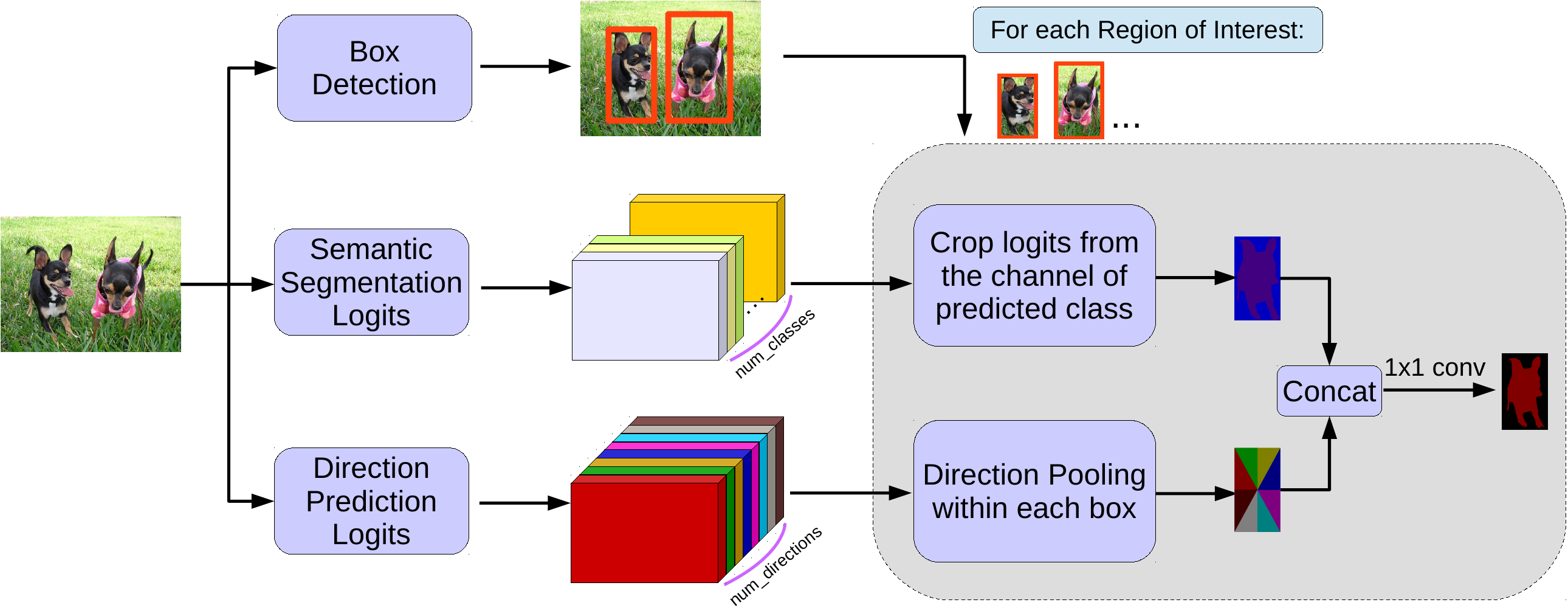} \\
  \end{tabular}
  \caption{MaskLab generates three outputs, including refined box predictions (from Faster-RCNN), semantic segmentation logits (logits for pixel-wise classification), and direction prediction logits (logits for predicting each pixel's direction toward its corresponding instance center). For each region of interest, we perform foreground/background segmentation by exploiting semantic segmentation and direction logits. Specifically, for the semantic segmentation logits, we pick the channel based on the predicted box label and crop the regions according to the predicted box. For the direction prediction logits, we perform the direction pooling to assemble the regional logits from each channel. These two cropped features are concatenated and passed through another $1\times1$ convolution for foreground/background segmentation.}
  \label{fig:model_overview}
\end{figure*}

Specifically, MaskLab builds on top of Faster R-CNN \cite{ren2015faster} and additionally produces two
outputs: semantic segmentation and instance center direction \cite{uhrig2016pixel}. The predicted boxes
returned by Faster R-CNN bring object instances of different scales to a canonical scale, and MaskLab
performs foreground/background segmentation within each predicted box by exploiting both semantic segmentation
and direction prediction. The semantic segmentation prediction, encoding the pixel-wise classification
information including \textit{background} class, is adopted to distinguish between objects of different
semantic classes (\eg, person and background), and thus removes the duplicate background encoding in \cite{dai2017fully}.
Additionally, direction prediction is used to separate object instances of the same
semantic label. Our model employs the same assembling operation in \cite{dai2016instancesensitive, dai2017fully}
to collect the direction information and thus gets rid of the complicate template matching used in \cite{uhrig2016pixel}.
Furthermore, motivated by the recent advances in both segmentation and detection, MaskLab further incorporates
atrous convolution \cite{chen2017deeplab} to extract denser features maps, hypercolumn features \cite{hariharan2014hypercolumns}
for refining mask segmentation \cite{eigen2015predicting}, multi-grid \cite{wang2017understanding, dai2017deformable, chen2017rethinking}
for capturing different scales of context, and a new TensorFlow operation \cite{abadi2016tensorflow}, deformable crop and resize,
inspired by the deformable pooling operation \cite{dai2017deformable}.

We demonstrate the effectiveness of the proposed model on the challenging COCO instance segmentation
benchmark \cite{lin2014microsoft}. Our proposed model, MaskLab, shows comparable performance with
other state-of-art models in terms of both mask segmentation (\eg, FCIS \cite{dai2017fully} and Mask R-CNN \cite{he2017mask})
and box detection (\eg, G-RMI \cite{huang2016speed} and TDM \cite{shrivastava2016beyond}). Finally, we elaborate on the
implementation details and provide detailed ablation studies of the proposed model.

\section{Related Work}
In this work, we categorize current instance segmentation methods based on deep neural networks into two types, depending on how the method approaches the problem by starting from either detection or segmentation modules. 

\textbf{Detection-based methods:} This type of methods exploits state-of-art detection models (\eg, Fast-RCNN \cite{girshick2015fast}, Faster-RCNN \cite{ren2015faster} or R-FCN \cite{dai2016rfcn}) to either classify mask regions or refine the predicted boxes to obtain masks. There have been several methods developed for mask proposals, including CPMC \cite{carreira2012cpmc}, MCG \cite{arbelaez2014multiscale}, DeepMask \cite{pinheiro2015learning}, SharpMask \cite{pinheiro2016learning}, and instance-sensitive FCNs \cite{dai2016instancesensitive}. Recently, Zhang and He \cite{zhang2017deep} propose a free-form deformation network to refine the mask proposals. Coupled with the mask proposals, SDS \cite{hariharan2014simultaneous, chen2015multi} and CFM \cite{dai2015convolutional} incorporate mask-region features to improve the classification accuracy, while \cite{hariharan2014hypercolumns} exploit hypercolumn features (\ie, features from the intermediate layers). Li \etal \cite{li2016iterative} iteratively apply the prediction. Zagoruyko \etal \cite{Zagoruyko2016Multipath} exploit object context at multiple scales. The work of MNC \cite{dai2016instance} shows promising results by decomposing instance segmentation into three sub-problems including box localization, mask refinement and instance classification. Hayer \etal \cite{hayder2017bounary} improve MNC by recovering the mask boundary error resulted from box prediction. Arnab \etal \cite{arnab2016higher, arnab2017pixelwise} apply higher-order Conditional Random Fields (CRFs) to refine the mask results. FCIS \cite{dai2017fully}, the first Fully Convolutional Network (FCN) \cite{long2014fully} for instance segmentation, enriches the position-sensitive score maps from \cite{dai2016instancesensitive} by further considering inside/outside score maps. Mask-RCNN \cite{he2017mask}, built on top of FPN \cite{lin2016feature}, adds another branch to obtain refined mask results from Faster-RCNN box prediction and demonstrates outstanding performance.

\begin{figure*}[!t]
  \centering
  \begin{tabular}{c}
    \includegraphics[width=0.95\linewidth]{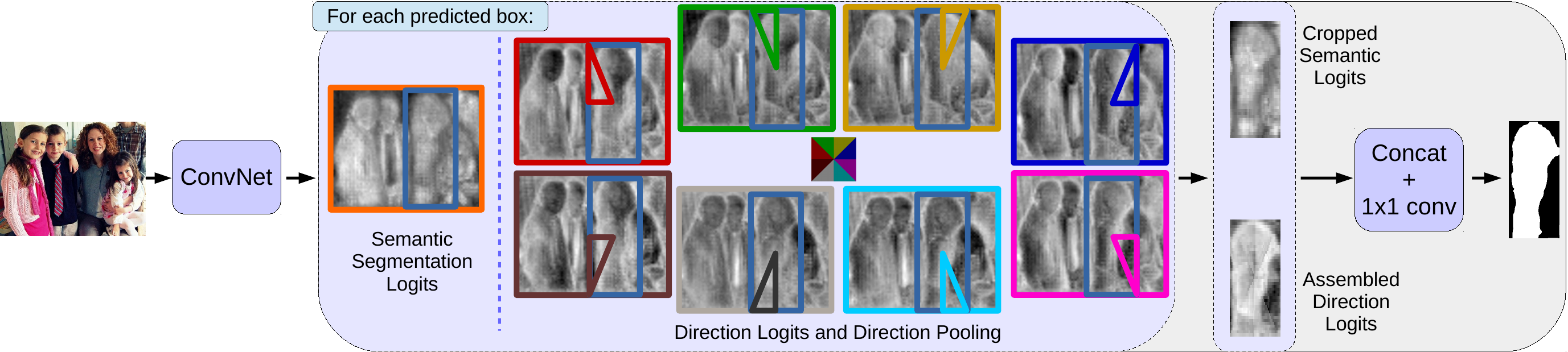} \\
  \end{tabular}
  \caption{Semantic segmentation logits and direction prediction logits are used to perform foreground/background segmentation within each predicted box. In particular, segmentation logits are able to distinguish between instances of different semantic classes (\eg, person and background), while direction logits (directions are color-coded) further separate instances of the same semantic class (\eg, two persons in the predicted blue box). In the assembling operation, regional logits (the color triangular regions) are copied from each direction channel, similar to \cite{dai2016instancesensitive, dai2017fully}. For example, the region specified by the red triangle copies the logits from the red direction channel encoding instance direction from 0 degree to 45 degree. Note the weak activations in the pink channel encoding instance direction from 180 degree to 225 degree.}
  \label{fig:model_illustration}
\end{figure*}

\textbf{Segmentation-based methods:} This type of methods generally adopt a two-stage processing, including segmentation and clustering. Pixel-level predictions are obtained by the segmentation module before the clustering process is applied to group them together for each object instance. Proposal-free network \cite{liang2015proposal} applies spectral clustering to group segmentation results from DeepLab \cite{chen2014semantic}, while Zhang \etal \cite{zhang2015monocular} exploit depth ordering within an image patch. In addition to semantic and depth information, Uhrig \etal \cite{uhrig2016pixel} further train an FCN to predict instance center direction. Zhang \etal \cite{zhang2016instance} propose a novel fully connected CRF \cite{krahenbuhl2011efficient} (with fast inference by permutohedral lattice \cite{adams2010fast}) to refine the results. Liu \etal \cite{liu2016multi} segment objects in multi-scale patches and aggregate the results. Levinkov \etal \cite{levinkov2017joint} propose efficient local search algorithms for instance segmentation. Wu \etal \cite{wu2016bridging} exploit a localization network for grouping, while Bai and Urtasun \cite{bai2017deep} adopt a Watershed Transform Net. Furthermore, Liu \etal \cite{liu2017sgn} propose to sequentially solve the grouping problem and gradually compose object instances. \cite{kirillov2017instancecut, jin2016object} exploit boundary detection information, while \cite{newell2017associative, fathi2017semantic, de2017semantic} propose to cluster instances \wrt the learned embedding values.

  In addition to the two categories, there is other interesting work. For example, \cite{romera2016recurrent, ren2017end} propose recurrent neural networks to sequentially segment an instance at a time. \cite{khoreva2016simple} propose a weakly supervised instance segmentation model given only bounding box annotations.

Our proposed MaskLab model combines the advantages from both detection-based and segmentation-based methods. In particular, MaskLab builds on top of Faster-RCNN \cite{ren2015faster} and additionally incorporates semantic segmentation (to distinguish between instances of different semantic classes, including background class) and direction features \cite{uhrig2016pixel} (to separate instances of the same semantic label). Our work is most similar to FCIS \cite{dai2017fully}, Mask R-CNN \cite{he2017mask}, and the work of \cite{uhrig2016pixel}; we build on top of Faster R-CNN \cite{ren2015faster} instead of R-FCN \cite{dai2016rfcn} (and thus replace the complicated template matching for instance detection in \cite{uhrig2016pixel}), exploit semantic segmentation prediction to remove duplicate background encoding in the inside/outside score maps, and we also simplify the position-sensitive pooling to direction pooling.

  

\section{MaskLab}

\textbf{Overview:} Our proposed model, MaskLab, employs ResNet-101 \cite{he2015deep} as feature extractor. It consists of three components with all features shared up to conv4 (or res4x) block and one extra duplicate conv5 (or res5x) block is used for the box classifier in Faster-RCNN \cite{ren2015faster}. Note that the original conv5 block is shared for both semantic segmentation and direction prediction. As shown in \figref{fig:model_overview}, MaskLab, built on top of Faster-RCNN \cite{ren2015faster}, produces box prediction (in particular, refined boxes after the box classifier), semantic segmentation logits (logits for pixel-wise classification) and direction prediction logits (logits for predicting each pixel's direction towards its corresponding instance center \cite{uhrig2016pixel}). Semantic segmentation logits and direction prediction logits are computed by another $1\times1$ convolution added after the last feature map in the conv5 block of ResNet-101. 
Given each predicted box (or region of interest), we perform foreground/background segmentation by exploiting those two logits. Specifically, we apply a class-agnostic (\ie, with weights shared across all classes) $1\times1$ convolution on the concatenation of (1) cropped semantic logits from the semantic channel predicted by Faster-RCNN and (2) cropped direction logits after direction pooling.

\textbf{Semantic and direction features:} MaskLab generates semantic segmentation logits and direction prediction logits for an image. The semantic segmentation logits are used to predict pixel-wise semantic labels, which are able to separate instances of different semantic labels, including the background class. On the other hand, the direction prediction logits are used to predict each pixel's direction towards its corresponding instance center and thus they are useful to further separate instances of the same semantic labels.

Given the predicted boxes and labels from the box prediction branch, we first select the channel \wrt the predicted label (\eg, the person channel) from the semantic segmentation logits, and crop the regions \wrt the predicted box. In order to exploit the direction information, we perform the same assembling operation in \cite{dai2016instancesensitive, dai2017fully} to gather regional logits (specified by the direction) from each direction channel. The cropped semantic segmentation logits along with the pooled direction logits are then used for foreground/background segmentation. We illustrate the details in \figref{fig:model_illustration}, which shows that the segmentation logits for `person` clearly separate the person class from background and the tie class, and the direction logits are able to predict the pixel's direction towards its instance center. After assembling the direction logits, the model is able to further separate the two persons within the specified box region. Note that our proposed direction prediction logits are class-agnostic instead of having the logits for each semantic class as in FCIS \cite{dai2017fully}, yielding more compact models. Specifically, for mask segmentation with $K$ classes, our model requires $(K + 32)$ channels ($K$ for semantic segmentation and 32 for direction pooling), while \cite{dai2017fully} outputs $2\times (K + 1) \times 49$ channels (2 for inside/outside score maps and 49 for position grids).

\begin{figure}[!t]
  \centering
  \begin{tabular}{c c c}
    \includegraphics[width=0.99\linewidth]{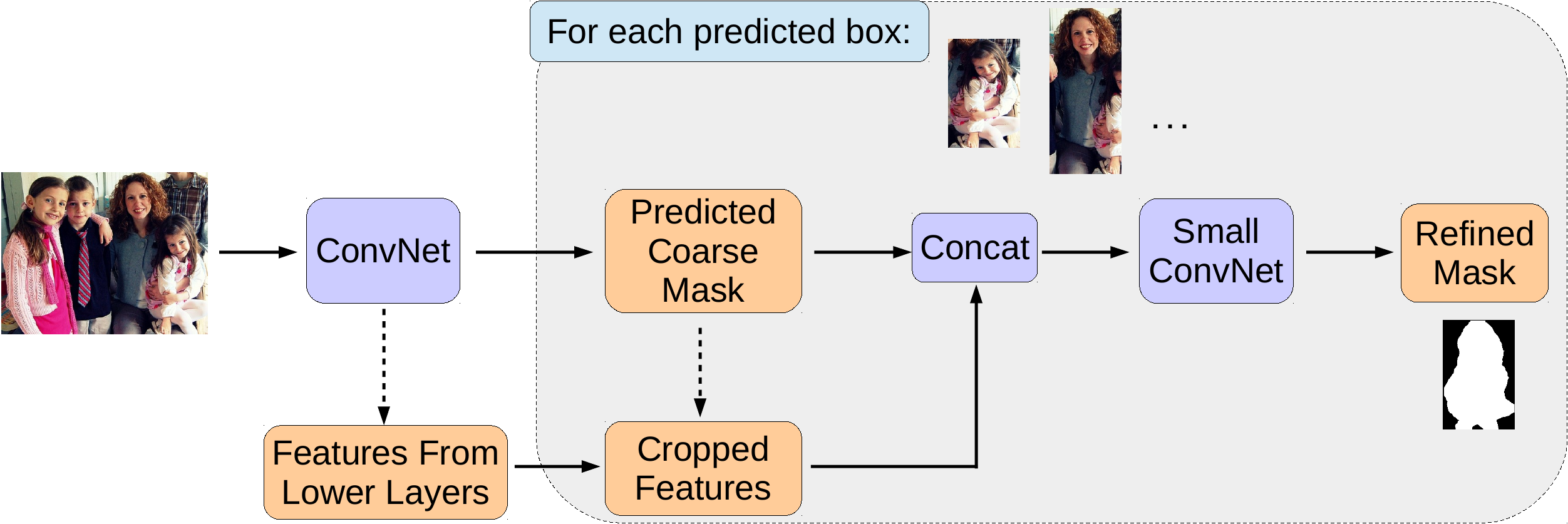} \\
  \end{tabular}
  \caption{Mask refinement. Hypercolumn features are concatenated with the coarse predicted mask and then fed to another small ConvNet to produce the final refined mask predictions.}
  \label{fig:refine_mask}
\end{figure}

\textbf{Mask refinement:} Motivated by \cite{eigen2015predicting} which applies another network consisting of only few layers for segmentation refinement, we further refine the predicted coarse masks by exploiting the hypercolumn features \cite{hariharan2014hypercolumns}. Specifically, as shown in \figref{fig:refine_mask}, the generated coarse mask logits (by only exploiting semantic and direction features) are concatenated with features from lower layers of ResNet-101, which are then processed by three extra convolutional layers in order to predict the final mask.

\begin{figure}[!t]
  \centering
  \begin{tabular}{c c c}
    \includegraphics[width=0.3\linewidth]{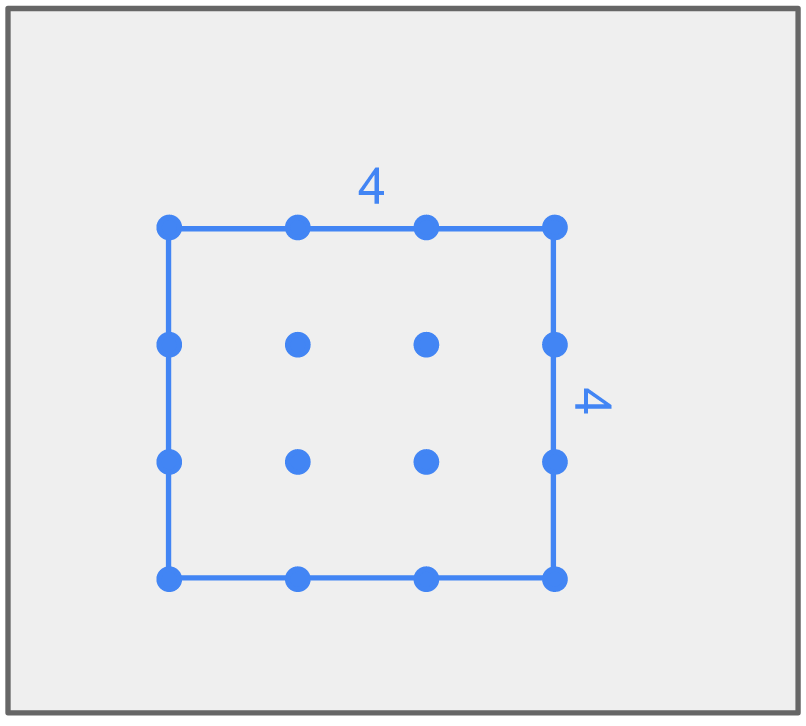} &
    \includegraphics[width=0.3\linewidth]{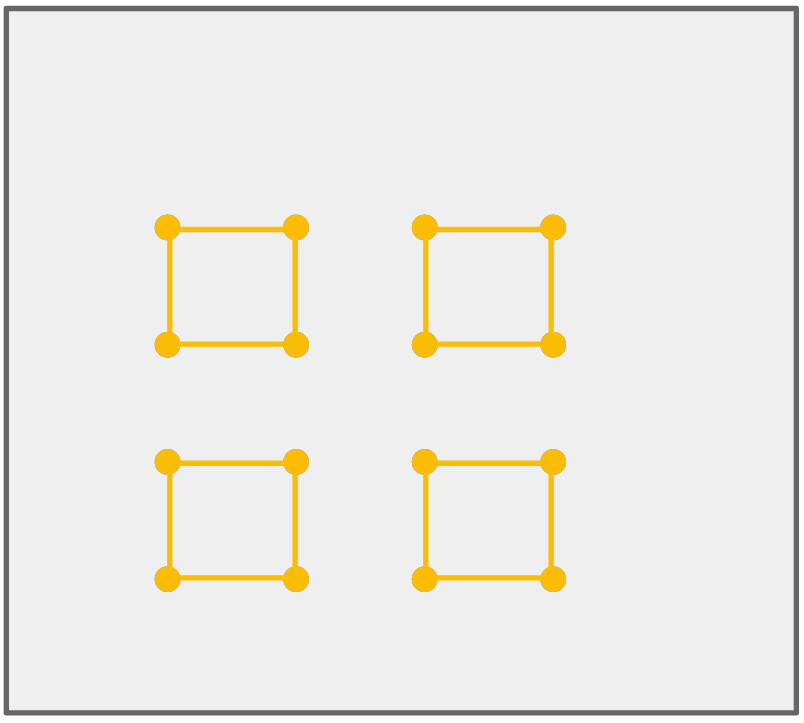} &
    \includegraphics[width=0.3\linewidth]{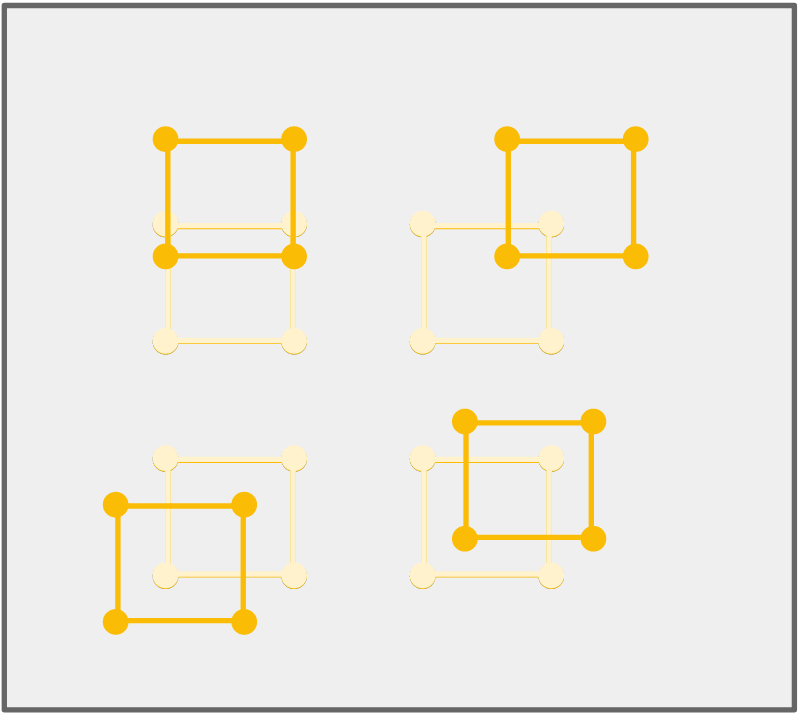} \\
    {\scriptsize (a) Crop and resize} &
    {\scriptsize (b) $2\times2$ sub-boxes} &
    {\scriptsize (c) Deformed sub-boxes} \\
  \end{tabular}
  \caption{Deformable crop and resize. (a) The operation, crop and resize, crops features within a bounding box region and resizes them to a specified size $4\times4$. (b) The $4\times4$ region is then divided into 4 small sub-boxes, and each has size $2\times2$. (c) Another small network is applied to learn the offsets of each sub-box. Then we perform crop and resize again \wrt to the deformed sub-boxes.}
  \label{fig:deformed_crop_and_resize}
\end{figure}

\textbf{Deformable crop and resize:} Following Dai \etal \cite{dai2017deformable}, who demonstrate significant improvement in object detection by deforming convolution and pooling operations, we modify the key TensorFlow operation used for box classification, ``crop and resize'' (similar to RoIAlign in Mask R-CNN \cite{he2017mask}), to support deformation as well. As shown in \figref{fig:deformed_crop_and_resize}, ``crop and resize'' first crops a specified bounding box region from the feature maps and then bilinearly resizes them to a specified size (\eg, $4\times4$). We further divide the regions into several sub-boxes (\eg, 4 sub-boxes and each has size $2\times2$) and employ another small network to learn the offsets for each sub-box. Finally, we perform ``crop and resize'' again \wrt each deformed sub-box. In summary, we use ``crop and resize'' twice to implement the deformable pooling in \cite{dai2017deformable}.

\section{Experimental Evaluation}
We conduct experiments on the COCO dataset \cite{lin2014microsoft}. Our proposed model is implemented in TensorFlow \cite{abadi2016tensorflow} on top of the object detection library developed by \cite{huang2016speed}.

\subsection{Implementation Details}
We employ the same hyper-parameter settings as in \cite{huang2016speed, sun2017revisiting}, and only discuss the main difference below.

\textbf{Atrous convolution:} We apply the atrous convolution \cite{holschneider1989real, giusti2013fast, sermanet2013overfeat, papandreou2014untangling}, which has been successfully explored in semantic segmentation \cite{chen2015attention, zhao2017pyramid, chen2017rethinking}, object detection \cite{dai2016rfcn, huang2016speed} and instance segmentation \cite{zhang2015monocular, dai2017fully}, to extract denser feature maps. Specifically, we extract features with $\emph{output\_stride}=8$ (\emph{output\_stride} denotes the ratio of input image spatial resolution to final output resolution).

\textbf{Weight initialization:} For the $1\times1$ convolution applied to the concatenation of semantic and direction features, we found that the training converges faster by initializing the convolution weights to be $(0.5, 1)$, putting a slightly larger weight on the direction features, which is more important in instance segmentation, as shown in the experimental results.

\textbf{Mask training:} During training, only groundtruth boxes are used to train the branches that predict semantic segmentation logits and direction logits, since direction logits may not align well with instance center if boxes are jittered. We employ sigmoid function to estimate both the coarse and refined mask results. Our proposed model is trained end-to-end without piecewise pretraining of each component.

\subsection{Quantitative Results}
We first report the ablation studies on a \textit{minival} set and then evaluate the best model on \textit{test-dev} set, with the metric mean average precision computed using mask IoU.

\textbf{Mask crop size:} The TensorFlow operation, ``crop and resize'', is used at least in two places: one for box classification and one for cropping semantic and direction features for foreground/background segmentation (another one for deformed sub-boxes if ``deformable crop and resize'' is used). In the former case, we use the same setting as in \cite{huang2016speed, sun2017revisiting}, while in the latter case, the crop size determines the mask segmentation resolution. Here, we experiment with the effect of using different crop size in \tabref{tab:mask_crop_size} and observe that using crop size more than 41 does not change the performance significantly and thus we use 41 throughout the experiments.

\begin{table}[!t]
  \centering
  \scalebox{1}{
  \begin{tabular}{c | c}
    \toprule[0.2em]
    Mask crop size & mAP@0.5 \\
    \toprule[0.2em]
    21 & 50.92\% \\
    41 & 51.29\% \\
    81 & 51.17\% \\
    161 & 51.36\% \\
    321 & 51.24\% \\
    \bottomrule[0.1em]
  \end{tabular}
  }
  \caption{Using crop size = 41 is sufficient for mask segmentation.}
  \label{tab:mask_crop_size}
\end{table}

\begin{figure}[!t]
  \centering
  \begin{tabular}{c c c}
    \includegraphics[width=0.3\linewidth]{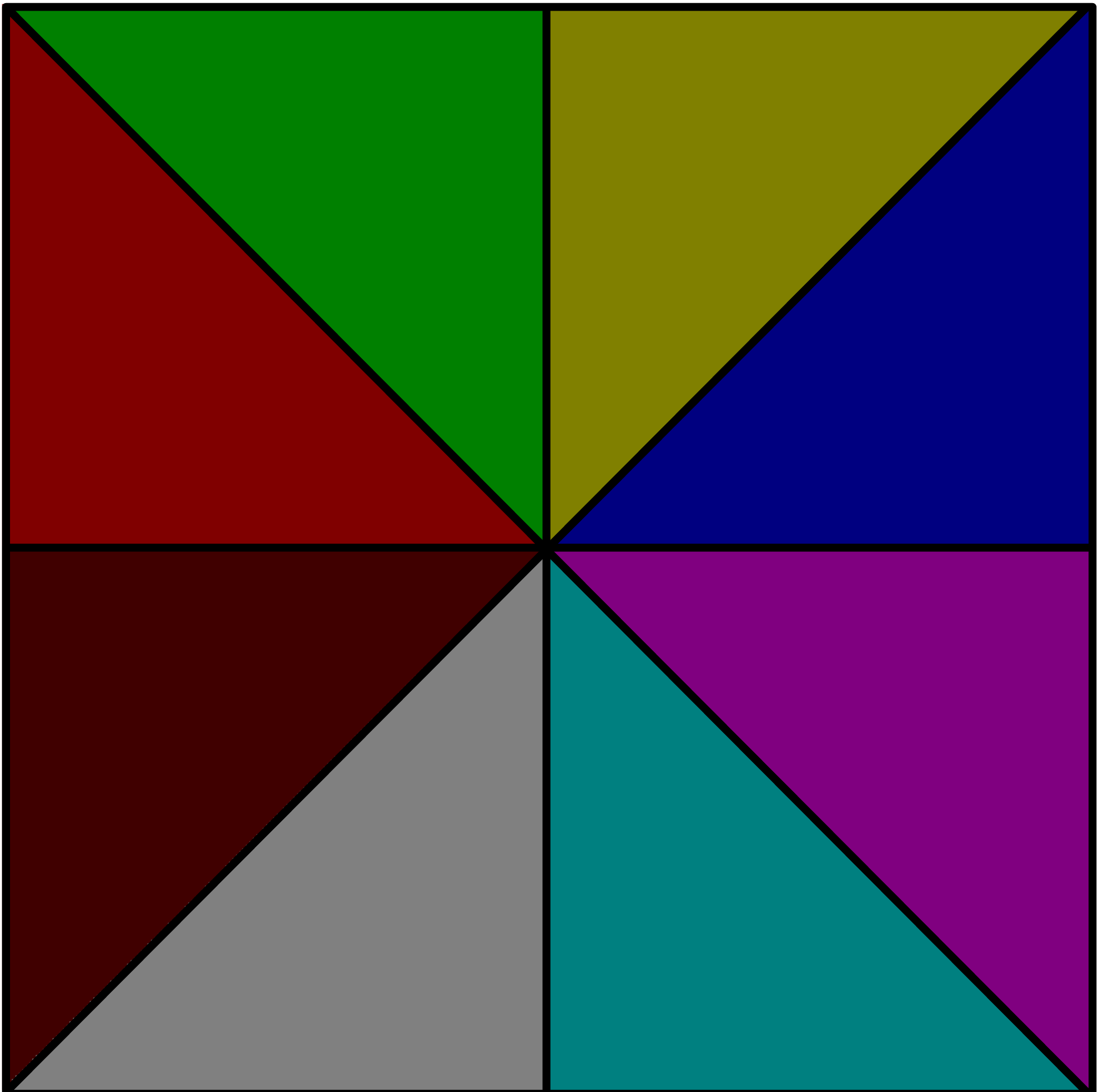} &
    \includegraphics[width=0.3\linewidth]{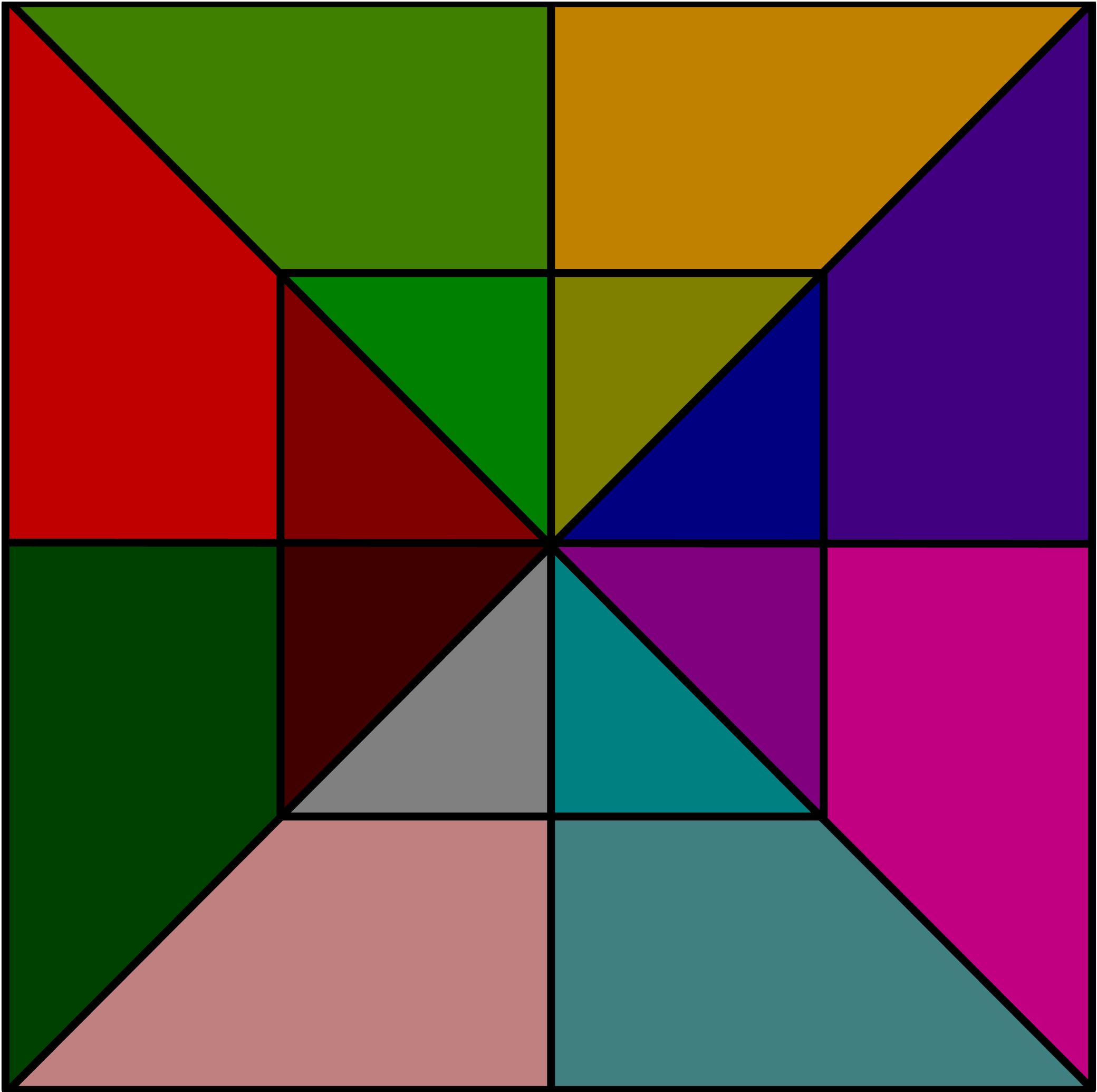} &
    \includegraphics[width=0.3\linewidth]{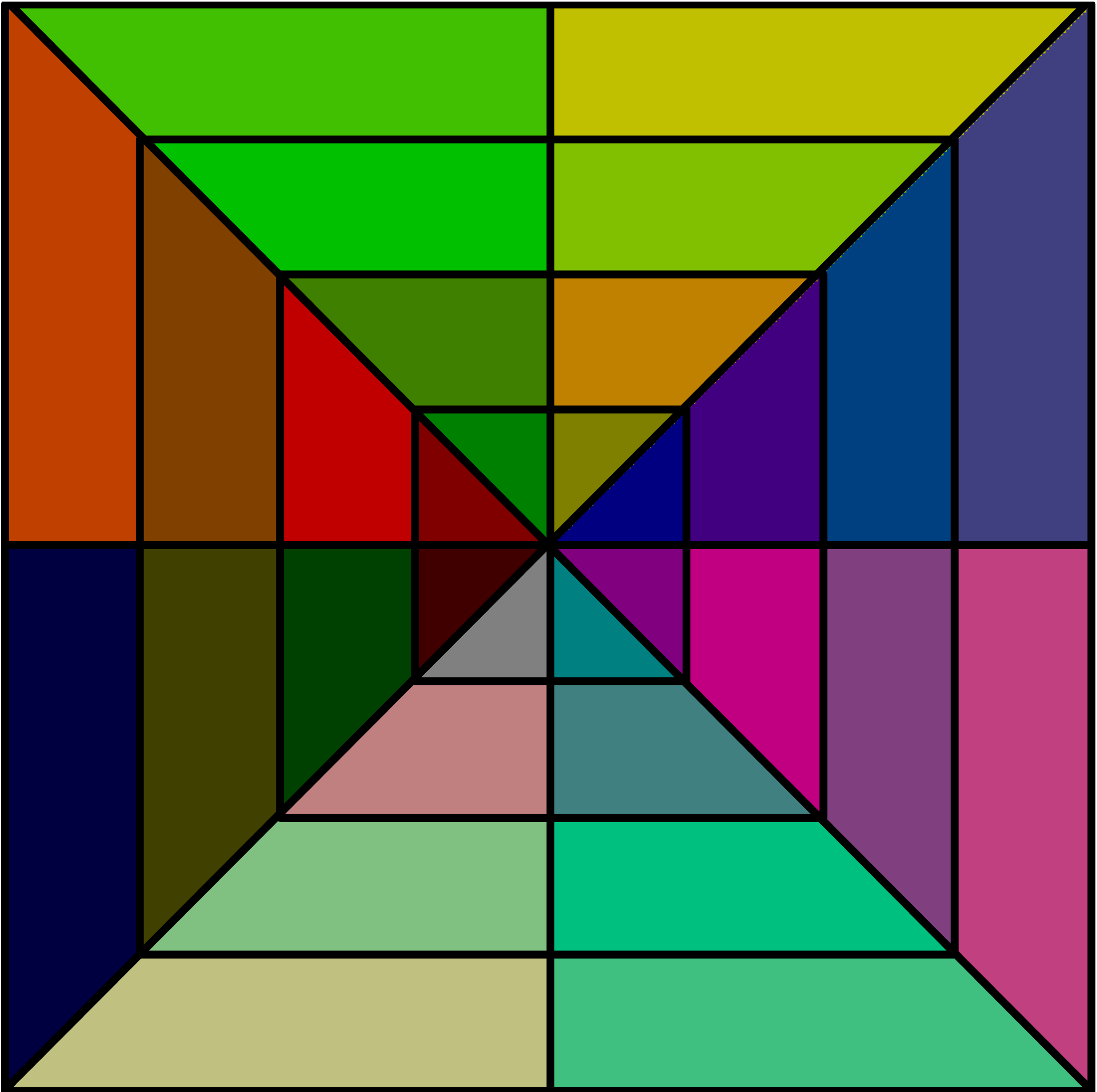} \\
    (a) 1 bin &
    (b) 2 bins &
    (c) 4 bins \\
  \end{tabular}
  \caption{We quantize the distance within each direction region. In (b), we split each original direction region into 2 regions. Our final model uses 4 bins for distance quantization as shown in (c).}
  \label{fig:boxes}
\end{figure}

\textbf{Effect of semantic and direction features:} In \tabref{tab:sem_dir}, we experiment with the effect of semantic and direction features. Given only semantic segmentation features, the model attains an mAP@0.75 performance of 24.44\%, while using only direction features the performance improves to 27.4\%, showing that direction feature is more important than the semantic segmentation feature. When employing both features, we achieve 29.72\%. We observe that the performance can be further improved if we also quantize the distance in the direction pooling. As illustrated in \figref{fig:boxes}, we also quantize the distance with different number of bins. For example, when using 2 bins, we split the same direction region into 2 regions. We found that using 4 bins can further improves performance to 30.57\%. Note that quantizing the distance bins improves more at high mAP threshold (\cf mAP@0.5 and mAP@0.75 in \tabref{tab:sem_dir}). In the case of using $x$ distance bins, the channels of direction logits become $8\times x$, since we use 8 directions by default (\ie, 360 degree is quantized into 8 directions). Thus, our model generates $32=8 \times 4$ channels for direction pooling in the end.

\begin{table}[!t]
  \centering
  \begin{tabular}{c c | c c}
    \toprule[0.2em]
    Semantic & Direction & mAP@0.5 & mAP@0.75 \\
    \toprule[0.2em]
    \checkmark &            & 48.41\% & 24.44\% \\
               & \checkmark (1) & 50.21\% & 27.40\% \\
    \checkmark & \checkmark (1) & 51.83\% & 29.72\% \\
    \toprule[0.1em]
    \checkmark & \checkmark (4) & {\bf 52.26\%} & {\bf 30.57\%} \\
    \bottomrule[0.1em]
  \end{tabular}
  \caption{Effect of semantic and direction features. Direction features are more important than semantic segmentation features in the model, and the best performance is obtained by using both features and adopting 4 bins to quantize the distance in direction pooling. We show number of bins for distance quantization in parentheses.}
  \label{tab:sem_dir}
\end{table}

\textbf{Number of directions:} In \tabref{tab:direction_effect}, we explore the effect of different numbers of directions for quantizing the 360 degree. We found that using 8 directions is sufficient to deliver good performance, when adopting 4 bins for distance quantization. Our model thus uses $32=8\times4$ (8 for direction and 4 for distance quantization) channels for direction pooling throughout the experiments.

\begin{table}[!t]
  \centering
  \scalebox{1}{
  \begin{tabular}{c c | c c}
    \toprule[0.2em]
    Distance bins & Directions & mAP@0.5 & mAP@0.75\\
    \toprule[0.2em]
    4 & 2 & 53.51\% & 33.80\% \\
    4 & 4 & 53.85\% & 34.39\% \\
    4 & 6 & 54.10\% & 34.86\% \\
    4 & 8 & 54.13\% & 34.82\% \\
    \bottomrule[0.1em]
  \end{tabular}
  }
  \caption{Effect of different numbers of directions (\ie, how many directions for quantizing the 360 degree) when using four bins for distance quantization.}
  \label{tab:direction_effect}
\end{table}



\textbf{Mask refinement:} We adopt a small ConvNet consisting of three $5\times5$ convolution layers with 64 filters. We have experimented with replacing the small ConvNet with other structures (\eg, more layers and more filters) but have not observed any significant difference. In \tabref{tab:refine_mask}, we experiment with different features from lower-level of ResNet-101. Using conv1 (the feature map generated by the first convolution) improves the mAP@0.75 performance to 32.92\% from 30.57\%, while using both conv1 and conv2 (\ie, the last feature map in res2x block) obtains the best performance of 33.89\%. We have observed no further improvement when adding more lower-level features.

\begin{table}[!t]
  \centering
  \scalebox{1}{
  \begin{tabular}{c c c | c c}
    \toprule[0.2em]
    conv1 & conv2 & conv3 & mAP@0.5 & mAP@0.75\\
    \toprule[0.2em]
    &        &        & 52.26\% & 30.57\% \\
    \checkmark &      &   & 52.68\% & 32.92\% \\
    \checkmark & \checkmark &  & \bf{53.26\%} & \bf{33.89\%} \\
    \checkmark & \checkmark & \checkmark & 52.55\% & 32.88\% \\ \bottomrule[0.1em]
  \end{tabular}
  }
  \caption{Mask refinement. The best performance is obtained when using features from conv1 and conv2 (\ie, last feature map in res2x block). Note conv3 denotes the last feature map in res3x block.}
  \label{tab:refine_mask}
\end{table}

\textbf{Multi-grid:} Motivated by the success of employing a hierarchy of different atrous rates in semantic segmentation \cite{wang2017understanding, dai2017deformable, chen2017rethinking}, we modify the atrous rates in (1) the last residual block shared for predicting both semantic and direction features, and (2) the block for box classifier. Note that there are only three convolutions in those blocks. As shown in \tabref{tab:multi_grid}, it is more effective to apply different atrous rates for the box classifier. We think current evaluation metric (mAP$^r$) favors detection-based methods (as also pointed out by \cite{bai2017deep}) and thus it is more effective to improve the detection branch over the segmentation branch in our proposed model.

\begin{table}[!t]
  \centering
  \scalebox{1.}{
  \begin{tabular}{c c | c c c}
    \toprule[0.2em]
    & & \multicolumn{3}{c}{Box Classifier} \\
    & & (1, 1, 1)  & (1, 2, 1) & (1, 2, 4)\\
    \toprule[0.1em]
    \multirow{3}{*}{Sem/Dir} &  (4, 4, 4) & 34.82\% & 35.59\% & 35.35\% \\
    &  (4, 8, 4) & 35.07\% & 35.60\% & \bf{35.78\%} \\
    &  (4, 8, 16) & 34.89\% & 35.43\% & 35.51\% \\
    \bottomrule[0.1em]
  \end{tabular}
  }
  \caption{Multi-grid performance (mAP@0.75). Within the parentheses, we show the three atrous rates used for the three convolutions in the residual block. It is effective to adopt different atrous rates for the box classifier. Further marginal improvement is obtained when we also change the atrous rates in the last block that is shared by semantic segmentation and direction prediction logits.}
  \label{tab:multi_grid}
\end{table}


\textbf{Pretrained network:} We experimentally found that it is beneficial to pretrain the network. Recall that we duplicate one extra conv5 (or res5x) block in original ResNet-101 for box classification. As shown in \tabref{tab:ablation}, initializing the box classifier in Faster R-CNN with the ImageNet pretrained weights improves the performance from 33.89\% to 34.82\% (mAP@0.75). If we further pretrain ResNet-101 on the COCO semantic segmentation annotations and employ it as feature extractor, the model yields about 1\% improvement. This finding bears a similarity to \cite{bell2016inside} which adopts the semantic segmentation regularizer.

\textbf{Putting everything together:} We then employ the best multi-grid setting from \tabref{tab:multi_grid} and observe about 0.7\% improvement (mAP@0.75) over the one pretrained with segmentation annotations, as shown in \tabref{tab:ablation}. Following \cite{lin2016feature, he2017mask}, if the input image is resized to have a shortest side of 800 pixels and the Region Proposal Network adopts 5 scales, we observe another 1\% improvement. Using the implemented ``deformable crop and resize'' brings extra 1\% improvement. Additionally, we employ scale augmentation, specifically random scaling of inputs during training (with shortest side randomly selected from $\{480, 576, 688, 800, 930\}$), and attain performance of 40.41\% (mAP@0.75). Finally, we exploit the model that has been pretrained on the JFT-300M dataset \cite{hinton2015distilling, chollet2016xception, sun2017revisiting}, containing 300M images and more than 375M noisy image-level labels, and achieve performance of 41.59\% (mAP@0.75). 

\textbf{Atrous convolution for denser feature maps:} We employ atrous convolution, a powerful tool to control output resolution, to extract denser feature maps with $\emph{output\_stride}=8$. We have observed that our performance drops from 40.41\% to 38.61\% (mAP@0.75), if we change $\emph{output\_stride}=16$.

\begin{table}[!t]
  \centering
  \scalebox{0.75}{
  \begin{tabular}{c c c c c c c | c c}
    \toprule[0.2em]
    \textbf{BC} & \textbf{Seg} & \textbf{MG} & \textbf{Anc} & \textbf{DC} & \textbf{RS} & \textbf{JFT} & mAP@0.5 & mAP@0.75\\
    \toprule[0.2em]
               &            &            &            &        &      &  & 53.26\% & 33.89\% \\
    \checkmark &            &            &            &        &      &  & 54.13\% & 34.82\% \\
    \checkmark & \checkmark &            &            &        &      &  & 55.03\% & 35.91\% \\
    \checkmark & \checkmark & \checkmark &            &        &      &  & 55.64\% & 36.65\% \\
    \checkmark & \checkmark & \checkmark & \checkmark &        &      &  & 57.44\% & 37.57\% \\
    \checkmark & \checkmark & \checkmark & \checkmark & \checkmark  &      &  & 58.69\% & 38.61\% \\
    \checkmark & \checkmark & \checkmark & \checkmark & \checkmark  &  \checkmark  &  & 60.55\% & 40.41\% \\
    \checkmark & \checkmark & \checkmark & \checkmark & \checkmark  &  \checkmark  & \checkmark & 61.80\% & 41.59\% \\
    \bottomrule[0.1em]
  \end{tabular}
  }
  \caption{\textbf{BC}: Initialize the Box Classifier branch with ImageNet pretrained model. \textbf{Seg}: Pretrain the whole model on COCO semantic segmentation annotations. \textbf{MG}: Employ multi-grid in last residual block. \textbf{Anc}: Use (800, 1200) and 5 anchors. \textbf{DC}: Adopt deformable crop and resize. \textbf{RS}: Randomly scale inputs during training. \textbf{JFT}: Further pretrain the model on JFT dataset.}
  \label{tab:ablation}
\end{table}


\begin{table*}[!t]
  \centering
  \scalebox{1.}{
  \begin{tabular}{c c | c c c | c c c}
    \toprule[0.2em]
    Method & Feature Extractor & mAP & mAP@0.5 & mAP@0.75 & mAP$_S$ & mAP$_M$ & mAP$_L$ \\
    \toprule[0.2em]
    FCIS \cite{dai2017fully} & ResNet-101 & 29.2\% & 49.5\% & - & - & - & - \\
    FCIS+++ \cite{dai2017fully} & ResNet-101 & 33.6\% & 54.5\% & - & - & - & - \\
    Mask R-CNN \cite{he2017mask} & ResNet-101 & 33.1\% & 54.9\% & 34.8\% & 12.1\% & 35.6\% & 51.1\% \\
    Mask R-CNN \cite{he2017mask} & ResNet-101-FPN & 35.7\% & 58.0\% & 37.8\% & 15.5\% & 38.1\% & 52.4\% \\
    Mask R-CNN \cite{he2017mask} & ResNeXt-101-FPN & 37.1\% & 60.0\% & 39.4\% & 16.9\% & 39.9\% & 53.5\% \\
    \toprule[0.1em]
    MaskLab  & ResNet-101 & 35.4\% & 57.4\% & 37.4\% & 16.9\% & 38.3\% & 49.2\% \\
    MaskLab+ & ResNet-101       & 37.3\% & 59.8\% & 39.6\% & 19.1\% & 40.5\% & 50.6\% \\
    MaskLab+ & ResNet-101 (JFT) & 38.1\% & 61.1\% & 40.4\% & 19.6\% & 41.6\% & 51.4\% \\
    \bottomrule[0.1em]
  \end{tabular}
  }
  \caption{Instance segmentation single model \textit{mask} mAP on COCO test-dev. \textbf{MaskLab+}: Employ scale augmentation during training.}
  \label{tab:test_dev_mask}
\end{table*}

\begin{table*}[!t]
  \centering
  \scalebox{1.}{
  \begin{tabular}{c c | c c c | c c c}
    \toprule[0.2em]
    Method & Feature Extractor & mAP & mAP@0.5 & mAP@0.75 & mAP$_S$ & mAP$_M$ & mAP$_L$ \\
    \toprule[0.2em]
    G-RMI \cite{huang2016speed} & Inception-ResNet-v2 & 34.7\% & 55.5\% & 36.7\% & 13.5\% & 38.1\% & 52.0\% \\
    TDM \cite{shrivastava2016beyond} & Inception-ResNet-v2 & 37.3\% & 57.8\% & 39.8\% & 17.1\% & 40.3\% & 52.1\% \\
    Mask R-CNN \cite{he2017mask} & ResNet-101-FPN & 38.2\% & 60.3\% & 41.7\% & 20.1\% & 41.1\% & 50.2\% \\
    Mask R-CNN \cite{he2017mask} & ResNeXt-101-FPN & 39.8\% & 62.3\% & 43.4\% & 22.1\% & 43.2\% & 51.2\% \\
    \toprule[0.1em]
    MaskLab  & ResNet-101 & 39.6\% & 60.2\% & 43.3\% & 21.2\% & 42.7\% & 52.4\% \\
    MaskLab+ & ResNet-101       & 41.9\% & 62.6\% & 46.0\% & 23.8\% & 45.5\% & 54.2\% \\
    MaskLab+ & ResNet-101 (JFT) & 43.0\% & 63.9\% & 47.1\% & 24.8\% & 46.7\% & 55.2\% \\
    \bottomrule[0.1em]
  \end{tabular}
  }
  \caption{Object detection single model \textit{box} mAP on COCO test-dev. \textbf{MaskLab+}: Employ scale augmentation during training.}
  \label{tab:test_dev_box}
\end{table*}

\textbf{Test-dev mask results:} After finalizing the design choices on the \textit{minival} set, we then evaluate our model on the \textit{test-dev} set. As shown in \tabref{tab:test_dev_mask}, our MaskLab model outperforms FCIS+++ \cite{dai2017fully}, although FCIS+++ employs scale augmentation and on-line hard example mining \cite{shrivastava2016training} during training as well as multi-scale processing and horizontal flip during test. Our ResNet-101 based model performs better than the ResNet-101 based Mask R-CNN \cite{he2017mask}, and attains similar performance as the ResNet-101-FPN based Mask R-CNN. Our ResNet-101 based model with scale augmentation during training, denoted as MaskLab+ in the table, performs 1.9\% better, attaining similar mAP with Mask R-CNN built on top of the more powerful ResNeXt-101-FPN \cite{lin2016feature, xie2017aggreated}. Furthermore, pretraining MaskLab+ on the JFT dataset achieves performance of 38.1\% mAP.

\textbf{Test-dev box results:} We also show box detection results on COCO test-dev in \tabref{tab:test_dev_box}. Our ResNet-101 based model even without scale augmentation during training performs better than G-RMI \cite{huang2016speed} and TDM \cite{shrivastava2016beyond} which employ more expensive yet powerful Inception-ResNet-v2 \cite{szegedy2017inception} as feature extractor. All our model variants perform comparably or better than Mask R-CNN variants in the box detection task. Our best single-model result is obtained with scale augmentation during training, 41.9\% mAP with an ImageNet pretrained network and 43.0\% mAP with a JFT pretrained network.

\subsection{Qualitative Results}

\begin{figure}[!t]
  \centering
  \begin{tabular}{c c}
    \includegraphics[width=0.45\linewidth]{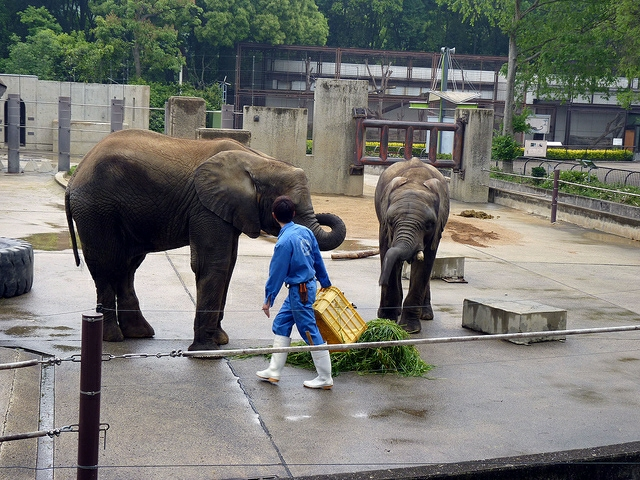} &
    \includegraphics[width=0.45\linewidth]{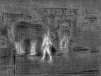} \\
    \includegraphics[width=0.45\linewidth]{fig/model_illustration/image/000000000458.png} &
    \includegraphics[width=0.45\linewidth]{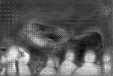} \\
    (a) Image &
    (b) `Person' Logits\\
  \end{tabular}
  \caption{`Person' channel in the predicted semantic segmentation logits. Note the high activations on non-person regions, since the semantic segmentation branch is only trained with groundtruth boxes. This, however, is being handled by the box detection branch which filters out wrong box predictions.}
  \label{fig:sem_seg_logits}
\end{figure}

\textbf{Semantic and direction features:} In \figref{fig:sem_seg_logits}, we visualize the `person' channel in the learned semantic segmentation logits. We have observed that there can be some high activations in the non-person regions (\eg, regions that are near elephant's legs and kite), since the semantic segmentation branch is only trained with groundtruth boxes without any negative ones. This, however, is being handled by the box detection branch which filters out wrong box predictions. More learned semantic segmentation and direction prediction logits are visualized in \figref{fig:model_illustration}.

\begin{figure}[!t]
  \centering
  \begin{tabular}{c c c}
    \includegraphics[height=0.22\linewidth]{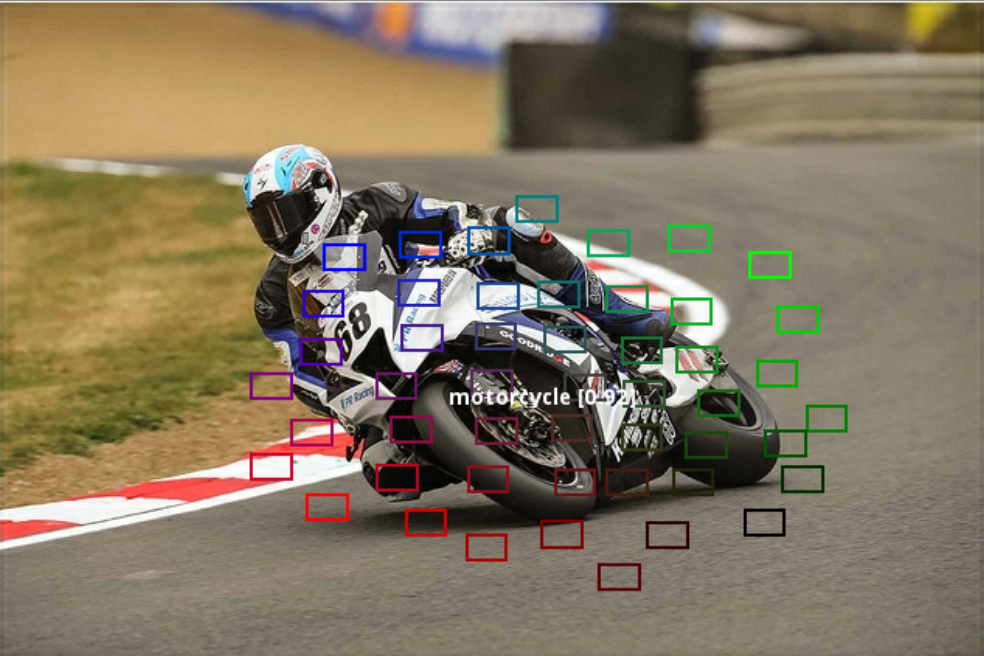} &
    \includegraphics[height=0.22\linewidth]{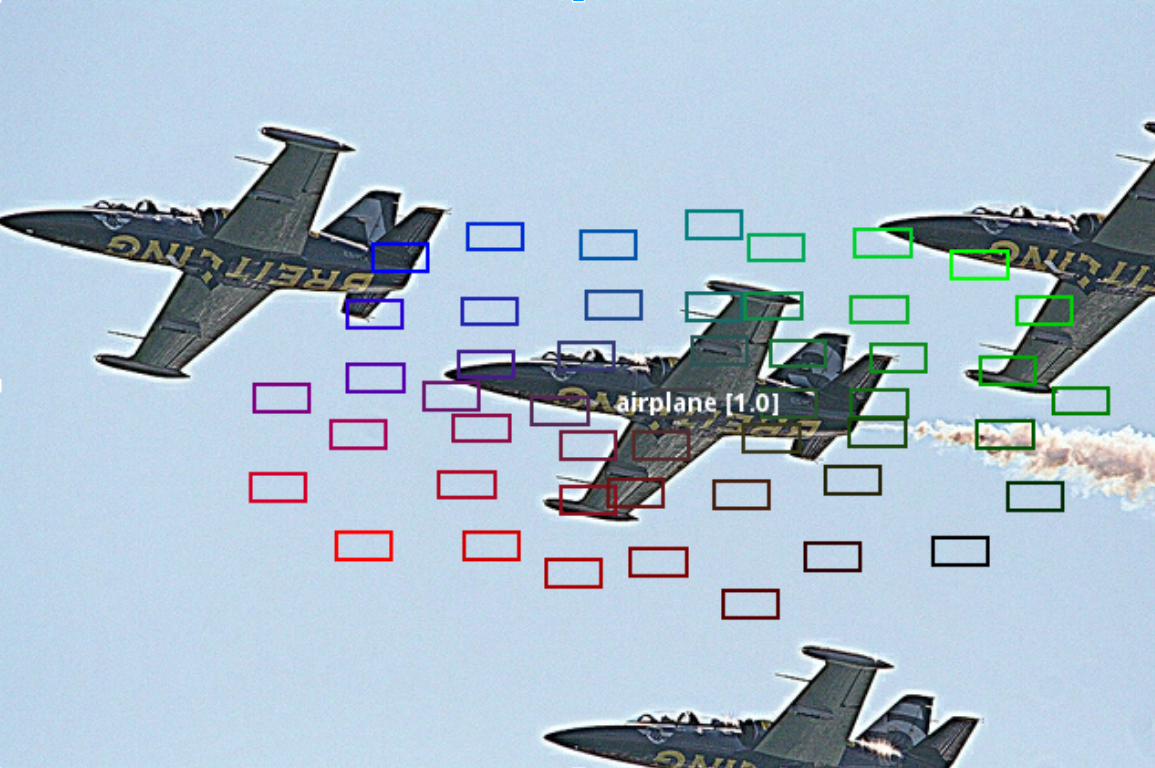} &
    \multicolumn{1}{|c}{
    \includegraphics[height=0.22\linewidth]{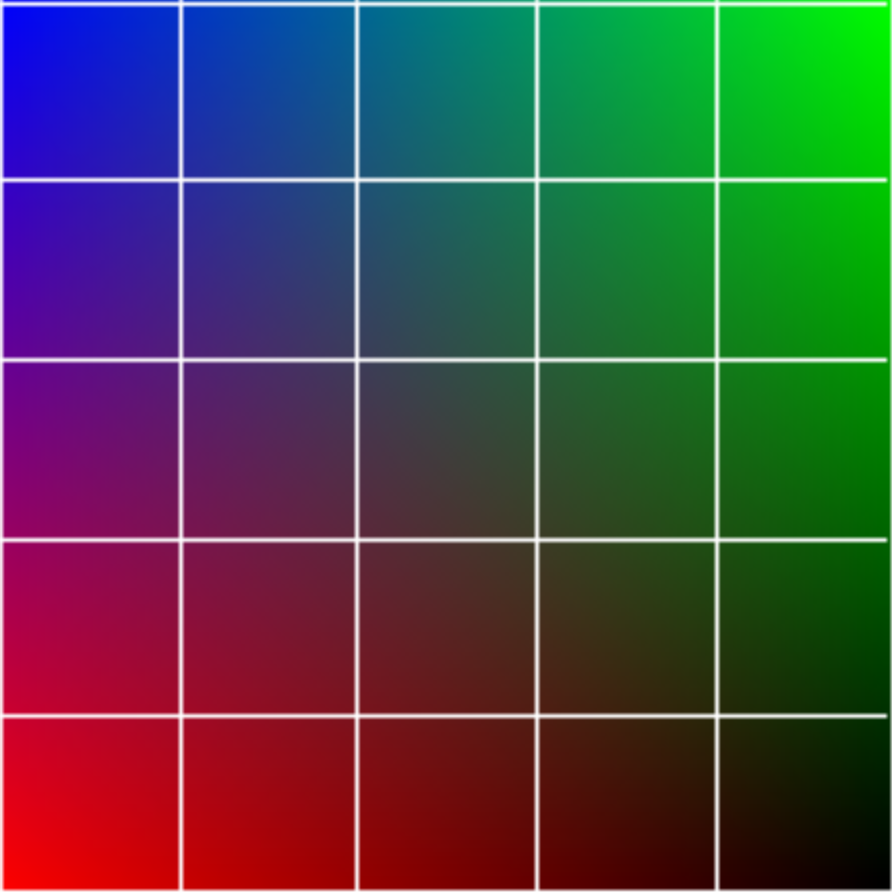}} \\
    \includegraphics[height=0.35\linewidth]{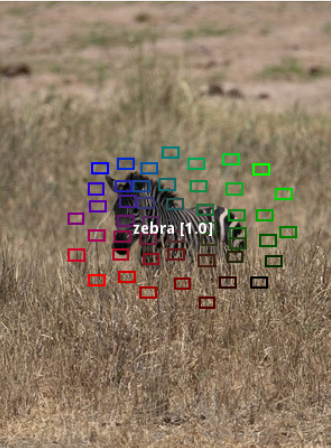} &
    \includegraphics[height=0.35\linewidth]{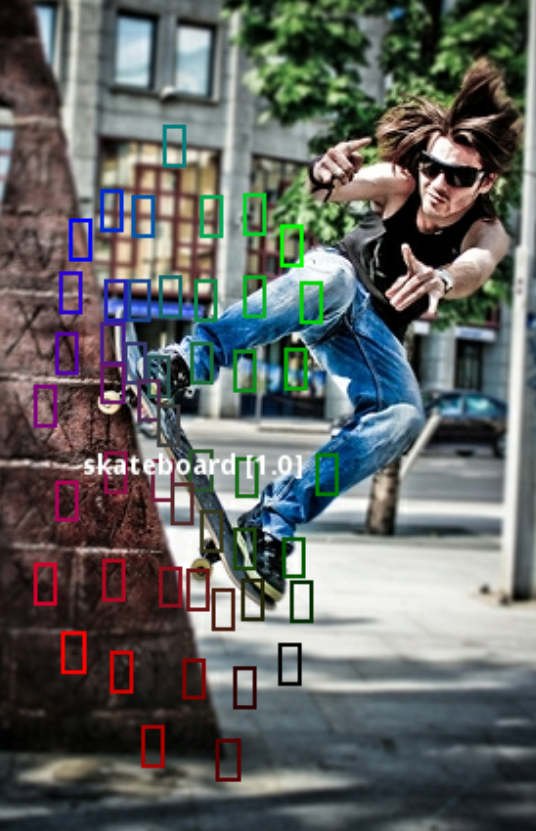} &
    \includegraphics[height=0.35\linewidth]{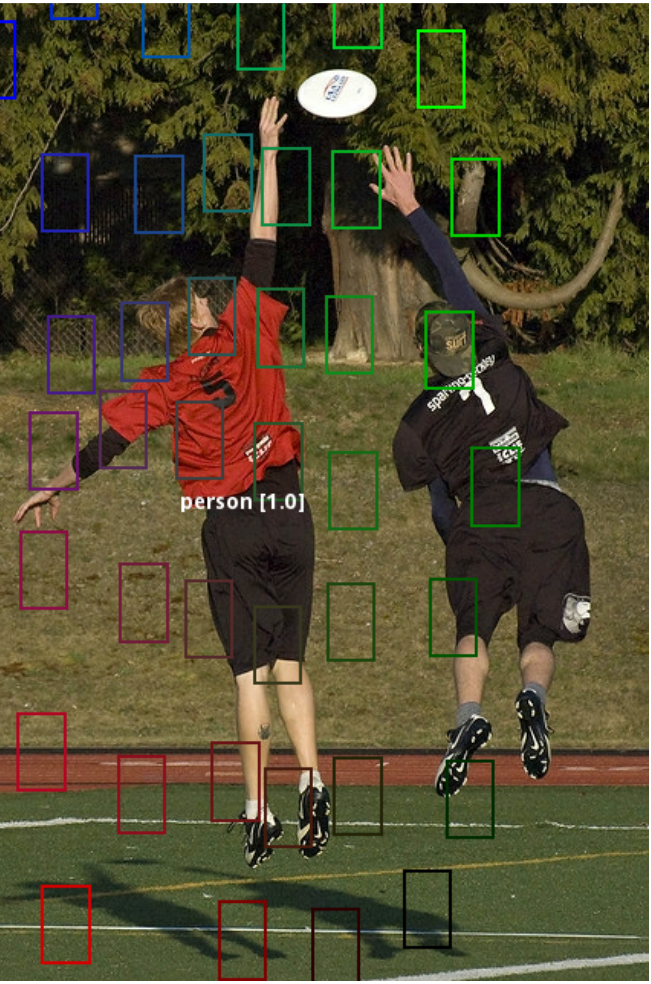} \\
  \end{tabular}
  \caption{Visualization of learned deformed sub-boxes. The 49 (arranged in a $7\times7$ grid) sub-boxes (each has size $2\times2$) are color-coded w.r.t. the top right panel (\eg, the top-left sub-box is represented by light blue color). Our ``deformable crop and resize'' tend to learn circle-shaped context for box classification.}
  \label{fig:vis_deformed_crop_and_resize}
\end{figure}

\textbf{Deformable crop and resize:} In \figref{fig:vis_deformed_crop_and_resize}, we visualize the learned deformed sub-boxes. Interestingly, unlike the visualization results of deformable pooling in \cite{dai2017deformable} which learns to focus on object parts, our sub-boxes are deformed in a circle-shaped arrangement, attempting to capture longer context for box classification. We note that incorporating context to improve detection performance has been used in, \eg, \cite{gidaris2015object, zhu2015segdeepm,Zagoruyko2016Multipath}, and our model is also able to learn this.

\begin{figure*}[!t]
  \centering
  \scalebox{1}{
  \begin{tabular}{c}
    \includegraphics[width=0.95\linewidth]{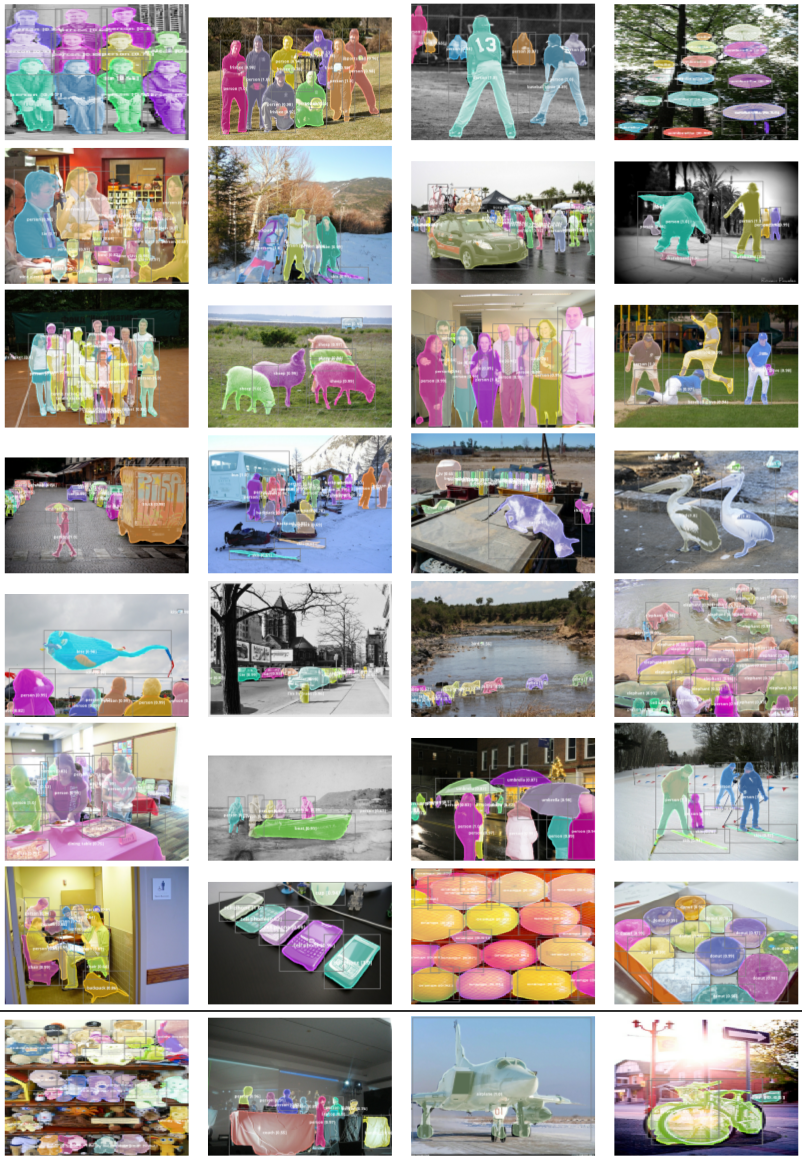} \\
  \end{tabular}
  }
  \caption{Visualization results on the \textit{minival} set. As shown in the figure (particularly, last row), our failure mode comes from two parts: (1) detection failure (missed-detection and wrong classification), and (2) failure to capture sharp object boundary.}
  \label{fig:vis_val}
\end{figure*}

\textbf{Predicted masks:} We show some qualitative results produced by our proposed model in \figref{fig:vis_val}. We further visualize our failure mode in the last row, mainly resulting from detection failure (\eg, missed-detection and wrong class prediction) and segmentation failure (\eg, coarse boundary result).

\section{Conclusion}
In this paper, we have presented a model, called MaskLab, that produces three outputs: box detection, semantic segmentation and direction prediction, for solving the problem of instance segmentation. MaskLab, building on top of state-of-art detector, performs foreground/background segmentation by utilizing semantic segmentation and direction prediction. We have demonstrated the effectiveness of MaskLab on the challenging COCO instance segmentation benchmark and shown promising results.

\ifcvprfinal
\paragraph{Acknowledgments}
We would like to acknowledge valuable discussions with Xuming He and Chen Sun, comments from Menglong Zhu and Xiao Zhang, and the support from the Google Mobile Vision team.
\fi

{\small
\bibliographystyle{ieee}
\bibliography{egbib}

\begin{thebibliography}{10}\itemsep=-1pt

\bibitem{abadi2016tensorflow}
M.~Abadi, A.~Agarwal, et~al.
\newblock Tensorflow: Large-scale machine learning on heterogeneous distributed
  systems.
\newblock {\em arXiv:1603.04467}, 2016.

\bibitem{adams2010fast}
A.~Adams, J.~Baek, and M.~A. Davis.
\newblock Fast high-dimensional filtering using the permutohedral lattice.
\newblock In {\em Eurographics}, 2010.

\bibitem{arbelaez2014multiscale}
P.~Arbel{\'a}ez, J.~Pont-Tuset, J.~T. Barron, F.~Marques, and J.~Malik.
\newblock Multiscale combinatorial grouping.
\newblock In {\em CVPR}, 2014.

\bibitem{arnab2016higher}
A.~Arnab, S.~Jayasumana, S.~Zheng, and P.~Torr.
\newblock Higher order conditional random fields in deep neural networks.
\newblock In {\em ECCV}, 2016.

\bibitem{arnab2017pixelwise}
A.~Arnab and P.~Torr.
\newblock Pixelwise instance segmentation with a dynamically instantiated
  network.
\newblock In {\em CVPR}, 2017.

\bibitem{bai2017deep}
M.~Bai and R.~Urtasun.
\newblock Deep watershed transform for instance segmentation.
\newblock In {\em CVPR}, 2017.

\bibitem{bell2016inside}
S.~Bell, C.~L. Zitnick, K.~Bala, and R.~Girshick.
\newblock Inside-outside net: Detecting objects in context with skip pooling
  and recurrent neural networks.
\newblock In {\em CVPR}, 2016.

\bibitem{de2017semantic}
B.~D. Brabandere, D.~Neven, and L.~V. Gool.
\newblock Semantic instance segmentation with a discriminative loss function.
\newblock {\em arXiv:1708.02551}, 2017.

\bibitem{carreira2012cpmc}
J.~Carreira and C.~Sminchisescu.
\newblock {CPMC}: Automatic object segmentation using constrained parametric
  min-cuts.
\newblock {\em PAMI}, 34(7):1312--1328, 2012.

\bibitem{chen2014semantic}
L.-C. Chen, G.~Papandreou, I.~Kokkinos, K.~Murphy, and A.~L. Yuille.
\newblock Semantic image segmentation with deep convolutional nets and fully
  connected crfs.
\newblock In {\em ICLR}, 2015.

\bibitem{chen2017deeplab}
L.-C. Chen, G.~Papandreou, I.~Kokkinos, K.~Murphy, and A.~L. Yuille.
\newblock Deeplab: Semantic image segmentation with deep convolutional nets,
  atrous convolution, and fully connected crfs.
\newblock {\em TPAMI}, 2017.

\bibitem{chen2017rethinking}
L.-C. Chen, G.~Papandreou, F.~Schroff, and H.~Adam.
\newblock Rethinking atrous convolution for semantic image segmentation.
\newblock {\em arXiv:1706.05587}, 2017.

\bibitem{chen2015attention}
L.-C. Chen, Y.~Yang, J.~Wang, W.~Xu, and A.~L. Yuille.
\newblock Attention to scale: Scale-aware semantic image segmentation.
\newblock In {\em CVPR}, 2016.

\bibitem{chen2015multi}
Y.-T. Chen, X.~Liu, and M.-H. Yang.
\newblock Multi-instance object segmentation with occlusion handling.
\newblock In {\em CVPR}, 2015.

\bibitem{chollet2016xception}
F.~Chollet.
\newblock Xception: Deep learning with depthwise separable convolutions.
\newblock {\em arXiv:1610.02357}, 2016.

\bibitem{dai2016instancesensitive}
J.~Dai, K.~He, Y.~Li, S.~Ren, and J.~Sun.
\newblock Instance-sensitive fully convolutional networks.
\newblock In {\em ECCV}, 2016.

\bibitem{dai2015convolutional}
J.~Dai, K.~He, and J.~Sun.
\newblock Convolutional feature masking for joint object and stuff
  segmentation.
\newblock In {\em CVPR}, 2015.

\bibitem{dai2016instance}
J.~Dai, K.~He, and J.~Sun.
\newblock Instance-aware semantic segmentation via multi-task network cascades.
\newblock In {\em CVPR}, 2016.

\bibitem{dai2016rfcn}
J.~Dai, Y.~Li, K.~He, and J.~Sun.
\newblock R-fcn: Object detection via region-based fully convolutional
  networks.
\newblock In {\em NIPS}, 2016.

\bibitem{dai2017deformable}
J.~Dai, H.~Qi, Y.~Xiong, Y.~Li, G.~Zhang, H.~Hu, and Y.~Wei.
\newblock Deformable convolutional networks.
\newblock In {\em ICCV}, 2017.

\bibitem{eigen2015predicting}
D.~Eigen and R.~Fergus.
\newblock Predicting depth, surface normals and semantic labels with a common
  multi-scale convolutional architecture.
\newblock In {\em ICCV}, 2015.

\bibitem{erhan2014scalable}
D.~Erhan, C.~Szegedy, A.~Toshev, and D.~Anguelov.
\newblock Scalable object detection using deep neural networks.
\newblock In {\em CVPR}, 2014.

\bibitem{fathi2017semantic}
A.~Fathi, Z.~Wojna, V.~Rathod, P.~Wang, H.~O. Song, S.~Guadarrama, and K.~P.
  Murphy.
\newblock Semantic instance segmentation via deep metric learning.
\newblock {\em arXiv:1703.10277}, 2017.

\bibitem{gidaris2015object}
S.~Gidaris and N.~Komodakis.
\newblock Object detection via a multi-region and semantic segmentation-aware
  cnn model.
\newblock In {\em ICCV}, 2015.

\bibitem{girshick2015fast}
R.~Girshick.
\newblock Fast r-cnn.
\newblock In {\em ICCV}, 2015.

\bibitem{girshick2014rcnn}
R.~Girshick, J.~Donahue, T.~Darrell, and J.~Malik.
\newblock Rich feature hierarchies for accurate object detection and semantic
  segmentation.
\newblock In {\em CVPR}, 2014.

\bibitem{giusti2013fast}
A.~Giusti, D.~Ciresan, J.~Masci, L.~Gambardella, and J.~Schmidhuber.
\newblock Fast image scanning with deep max-pooling convolutional neural
  networks.
\newblock In {\em ICIP}, 2013.

\bibitem{hariharan2014simultaneous}
B.~Hariharan, P.~Arbel{\'a}ez, R.~Girshick, and J.~Malik.
\newblock Simultaneous detection and segmentation.
\newblock In {\em ECCV}, 2014.

\bibitem{hariharan2014hypercolumns}
B.~Hariharan, P.~Arbel{\'a}ez, R.~Girshick, and J.~Malik.
\newblock Hypercolumns for object segmentation and fine-grained localization.
\newblock In {\em CVPR}, 2015.

\bibitem{hayder2017bounary}
Z.~Hayder, X.~He, and M.~Salzmann.
\newblock Boundary-aware instance segmentation.
\newblock In {\em CVPR}, 2017.

\bibitem{he2017mask}
K.~He, G.~Gkioxari, P.~Doll{\'a}r, and R.~Girshick.
\newblock Mask r-cnn.
\newblock In {\em ICCV}, 2017.

\bibitem{he2015deep}
K.~He, X.~Zhang, S.~Ren, and J.~Sun.
\newblock Deep residual learning for image recognition.
\newblock {\em arXiv:1512.03385}, 2015.

\bibitem{hinton2015distilling}
G.~Hinton, O.~Vinyals, and J.~Dean.
\newblock Distilling the knowledge in a neural network.
\newblock In {\em NIPS}, 2014.

\bibitem{holschneider1989real}
M.~Holschneider, R.~Kronland-Martinet, J.~Morlet, and P.~Tchamitchian.
\newblock A real-time algorithm for signal analysis with the help of the
  wavelet transform.
\newblock In {\em Wavelets: Time-Frequency Methods and Phase Space}, pages
  289--297. 1989.

\bibitem{huang2016speed}
J.~Huang, V.~Rathod, C.~Sun, M.~Zhu, A.~Korattikara, A.~Fathi, I.~Fischer,
  Z.~Wojna, Y.~Song, S.~Guadarrama, and K.~Murphy.
\newblock Speed/accuracy trade-offs for modern convolutional object detectors.
\newblock In {\em CVPR}, 2017.

\bibitem{jin2016object}
L.~Jin, Z.~Chen, and Z.~Tu.
\newblock Object detection free instance segmentation with labeling
  transformations.
\newblock {\em arXiv:1611.08991}, 2016.

\bibitem{khoreva2016simple}
A.~Khoreva, R.~Benenson, J.~Hosang, M.~Hein, and B.~Schiele.
\newblock Simple does it: Weakly supervised instance and semantic segmentation.
\newblock In {\em CVPR}, 2017.

\bibitem{kirillov2017instancecut}
A.~Kirillov, E.~Levinkov, B.~Andres, B.~Savchynskyy, and C.~Rother.
\newblock Instancecut: from edges to instances with multicut.
\newblock In {\em CVPR}, 2017.

\bibitem{krahenbuhl2011efficient}
P.~Kr{\"a}henb{\"u}hl and V.~Koltun.
\newblock Efficient inference in fully connected crfs with gaussian edge
  potentials.
\newblock In {\em NIPS}, 2011.

\bibitem{krizhevsky2012imagenet}
A.~Krizhevsky, I.~Sutskever, and G.~E. Hinton.
\newblock Imagenet classification with deep convolutional neural networks.
\newblock In {\em NIPS}, 2012.

\bibitem{LeCun1998}
Y.~LeCun, L.~Bottou, Y.~Bengio, and P.~Haffner.
\newblock {Gradient-based learning applied to document recognition}.
\newblock In {\em Proc. IEEE}, 1998.

\bibitem{levinkov2017joint}
E.~Levinkov, J.~Uhrig, S.~Tang, M.~Omran, E.~Insafutdinov, A.~Kirillov,
  C.~Rother, T.~Brox, B.~Schiele, and B.~Andres.
\newblock Joint graph decomposition \& node labeling: Problem, algorithms,
  applications.
\newblock In {\em CVPR}, 2017.

\bibitem{li2016iterative}
K.~Li, B.~Hariharan, and J.~Malik.
\newblock Iterative instance segmentation.
\newblock In {\em CVPR}, 2016.

\bibitem{dai2017fully}
Y.~Li, H.~Qi, J.~Dai, X.~Ji, and Y.~Wei.
\newblock Fully convolutional instance-aware semantic segmentation.
\newblock In {\em CVPR}, 2017.

\bibitem{liang2015proposal}
X.~Liang, Y.~Wei, X.~Shen, J.~Yang, L.~Lin, and S.~Yan.
\newblock Proposal-free network for instance-level object segmentation.
\newblock {\em arXiv preprint arXiv:1509.02636}, 2015.

\bibitem{lin2015efficient}
G.~Lin, C.~Shen, A.~van~den Hengel, and I.~Reid.
\newblock Efficient piecewise training of deep structured models for semantic
  segmentation.
\newblock In {\em CVPR}, 2016.

\bibitem{lin2016feature}
T.-Y. Lin, P.~Doll{\'a}r, R.~Girshick, K.~He, B.~Hariharan, and S.~Belongie.
\newblock Feature pyramid networks for object detection.
\newblock In {\em CVPR}, 2017.

\bibitem{lin2014microsoft}
T.-Y. Lin et~al.
\newblock Microsoft {COCO}: Common objects in context.
\newblock In {\em ECCV}, 2014.

\bibitem{liu2017sgn}
S.~Liu, J.~Jia†, S.~Fidler, and R.~Urtasun.
\newblock Sgn: Sequential grouping networks for instance segmentation.
\newblock In {\em ICCV}, 2017.

\bibitem{liu2016multi}
S.~Liu, X.~Qi, J.~Shi, H.~Zhang, and J.~Jia.
\newblock Multi-scale patch aggregation (mpa) for simultaneous detection and
  segmentation.
\newblock In {\em CVPR}, 2016.

\bibitem{liu2015ssd}
W.~Liu, D.~Anguelov, D.~Erhan, C.~Szegedy, S.~Reed, C.-Y. Fu, and A.~C. Berg.
\newblock {SSD}: Single shot multibox detector.
\newblock In {\em ECCV}, 2016.

\bibitem{liu2015semantic}
Z.~Liu, X.~Li, P.~Luo, C.~C. Loy, and X.~Tang.
\newblock Semantic image segmentation via deep parsing network.
\newblock In {\em ICCV}, 2015.

\bibitem{long2014fully}
J.~Long, E.~Shelhamer, and T.~Darrell.
\newblock Fully convolutional networks for semantic segmentation.
\newblock In {\em CVPR}, 2015.

\bibitem{luo2017deep}
P.~Luo, G.~Wang, L.~Lin, and X.~Wang.
\newblock Deep dual learning for semantic image segmentation.
\newblock In {\em ICCV}, 2017.

\bibitem{newell2017associative}
A.~Newell and J.~Deng.
\newblock Associative embedding: End-to-end learning for joint detection and
  grouping.
\newblock In {\em NIPS}, 2017.

\bibitem{papandreou2015weakly}
G.~Papandreou, L.-C. Chen, K.~Murphy, and A.~L. Yuille.
\newblock Weakly- and semi-supervised learning of a dcnn for semantic image
  segmentation.
\newblock In {\em ICCV}, 2015.

\bibitem{papandreou2014untangling}
G.~Papandreou, I.~Kokkinos, and P.-A. Savalle.
\newblock Modeling local and global deformations in deep learning: Epitomic
  convolution, multiple instance learning, and sliding window detection.
\newblock In {\em CVPR}, 2015.

\bibitem{pinheiro2015learning}
P.~O. Pinheiro, R.~Collobert, and P.~Doll{\'a}r.
\newblock Learning to segment object candidates.
\newblock In {\em NIPS}, 2015.

\bibitem{pinheiro2016learning}
P.~O. Pinheiro, T.-Y. Lin, R.~Collobert, and P.~Doll{\'a}r.
\newblock Learning to refine object segments.
\newblock In {\em ECCV}, 2016.

\bibitem{redmon2016you}
J.~Redmon, S.~Divvala, R.~Girshick, and A.~Farhadi.
\newblock You only look once: Unified, real-time object detection.
\newblock In {\em CVPR}, 2016.

\bibitem{ren2017end}
M.~Ren and R.~S. Zemel.
\newblock End-to-end instance segmentation with recurrent attention.
\newblock In {\em CVPR}, 2017.

\bibitem{ren2015faster}
S.~Ren, K.~He, R.~Girshick, and J.~Sun.
\newblock Faster r-cnn: Towards real-time object detection with region proposal
  networks.
\newblock In {\em NIPS}, 2015.

\bibitem{romera2016recurrent}
B.~Romera-Paredes and P.~H.~S. Torr.
\newblock Recurrent instance segmentation.
\newblock In {\em ECCV}, 2016.

\bibitem{sermanet2013overfeat}
P.~Sermanet, D.~Eigen, X.~Zhang, M.~Mathieu, R.~Fergus, and Y.~LeCun.
\newblock Overfeat: Integrated recognition, localization and detection using
  convolutional networks.
\newblock In {\em ICLR}, 2014.

\bibitem{shrivastava2016training}
A.~Shrivastava, A.~Gupta, and R.~Girshick.
\newblock Training region-based object detectors with online hard example
  mining.
\newblock In {\em CVPR}, 2016.

\bibitem{shrivastava2016beyond}
A.~Shrivastava, R.~Sukthankar, J.~Malik, and A.~Gupta.
\newblock Beyond skip connections: Top-down modulation for object detection.
\newblock {\em arXiv:1612.06851}, 2016.

\bibitem{sun2017revisiting}
C.~Sun, A.~Shrivastava, S.~Singh, and A.~Gupta.
\newblock Revisiting unreasonable effectiveness of data in deep learning era.
\newblock In {\em ICCV}, 2017.

\bibitem{szegedy2017inception}
C.~Szegedy, S.~Ioffe, V.~Vanhoucke, and A.~Alemi.
\newblock Inception-v4, inception-resnet and the impact of residual connections
  on learning.
\newblock In {\em AAAI}, 2017.

\bibitem{szegedy2014scalable}
C.~Szegedy, S.~Reed, D.~Erhan, D.~Anguelov, and S.~Ioffe.
\newblock Scalable, high-quality object detection.
\newblock {\em arXiv:1412.1441}, 2014.

\bibitem{uhrig2016pixel}
J.~Uhrig, M.~Cordts, U.~Franke, and T.~Brox.
\newblock Pixel-level encoding and depth layering for instance-level semantic
  labeling.
\newblock {\em arXiv:1604.05096}, 2016.

\bibitem{wang2017understanding}
P.~Wang, P.~Chen, Y.~Yuan, D.~Liu, Z.~Huang, X.~Hou, and G.~Cottrell.
\newblock Understanding convolution for semantic segmentation.
\newblock {\em arXiv:1702.08502}, 2017.

\bibitem{wu2016bridging}
Z.~Wu, C.~Shen, and A.~van~den Hengel.
\newblock Bridging category-level and instance-level semantic image
  segmentation.
\newblock {\em arXiv:1605.06885}, 2016.

\bibitem{wu2016wider}
Z.~Wu, C.~Shen, and A.~van~den Hengel.
\newblock Wider or deeper: Revisiting the resnet model for visual recognition.
\newblock {\em arXiv:1611.10080}, 2016.

\bibitem{xie2017aggreated}
S.~Xie, R.~Girshick, P.~Dollár, Z.~Tu, and K.~He.
\newblock Aggregated residual transformations for deep neural networks.
\newblock In {\em CVPR}, 2017.

\bibitem{Zagoruyko2016Multipath}
S.~Zagoruyko, A.~Lerer, T.-Y. Lin, P.~O. Pinheiro, S.~Gross, S.~Chintala, and
  P.~Doll{\'{a}}r.
\newblock A multipath network for object detection.
\newblock In {\em BMVC}, 2016.

\bibitem{zhang2017deep}
H.~Zhang and X.~He.
\newblock Deep free-form deformation network for object-mask registration.
\newblock In {\em ICCV}, 2017.

\bibitem{zhang2016instance}
Z.~Zhang, S.~Fidler, and R.~Urtasun.
\newblock Instance-level segmentation for autonomous driving with deep densely
  connected mrfs.
\newblock In {\em CVPR}, 2016.

\bibitem{zhang2015monocular}
Z.~Zhang, A.~G. Schwing, S.~Fidler, and R.~Urtasun.
\newblock Monocular object instance segmentation and depth ordering with cnns.
\newblock In {\em ICCV}, 2015.

\bibitem{zhao2017pyramid}
H.~Zhao, J.~Shi, X.~Qi, X.~Wang, and J.~Jia.
\newblock Pyramid scene parsing network.
\newblock In {\em CVPR}, 2017.

\bibitem{zheng2015conditional}
S.~Zheng, S.~Jayasumana, B.~Romera-Paredes, V.~Vineet, Z.~Su, D.~Du, C.~Huang,
  and P.~Torr.
\newblock Conditional random fields as recurrent neural networks.
\newblock In {\em ICCV}, 2015.

\bibitem{zhu2015segdeepm}
Y.~Zhu, R.~Urtasun, R.~Salakhutdinov, and S.~Fidler.
\newblock segdeepm: Exploiting segmentation and context in deep neural networks
  for object detection.
\newblock In {\em CVPR}, 2015.

\end{thebibliography}
}

\end{document}